\documentclass[10pt,twocolumn,letterpaper]{article}

\usepackage{cvpr}
\usepackage{times}
\usepackage{epsfig}
\usepackage{graphicx}
\usepackage{subcaption}
\usepackage{amsmath}
\usepackage{amssymb}
\usepackage{multirow}
\usepackage[linesnumbered,ruled]{algorithm2e}
\usepackage{algpseudocode}
\usepackage{color}
\usepackage{cite}

\usepackage[breaklinks=true,bookmarks=false]{hyperref}

\cvprfinalcopy 

\def\cvprPaperID{****} 

\begin{document}

\title{Deep Decision Trees for Discriminative Dictionary Learning with Adversarial Multi-Agent Trajectories}

\author{Tharindu~Fernando \hspace{1cm} Sridha Sridharan \hspace{1cm} Clinton Fookes \hspace{1cm} Simon Denman\\
\\Image and Video Research Lab, SAIVT, Queensland University of Technology (QUT), Australia\\
{\tt\small  \{t.warnakulasuriya, s.denman, s.sridharan, c.fookes\}@qut.edu.au}
}

\maketitle

\begin{abstract}
   With the explosion in the availability of spatio-temporal tracking data in modern sports, there is an enormous opportunity to better analyse, learn and predict important events in adversarial group environments. In this paper, we propose a deep decision tree architecture for discriminative dictionary learning from adversarial multi-agent trajectories. We first build up a hierarchy for the tree structure by adding each layer and performing feature weight based clustering in the forward pass. We then fine tune the player role weights using back propagation. The hierarchical architecture ensures the interpretability and the integrity of the group representation. The resulting architecture is a decision tree, with leaf-nodes capturing a dictionary of multi-agent group interactions. Due to the ample volume of data available, we focus on soccer tracking data, although our approach can be used in any adversarial multi-agent domain. We present applications of proposed method for simulating soccer games as well as evaluating and quantifying team strategies.
\end{abstract}

\section{Introduction}
Coinciding with the advancement of computer vision based tracking technologies, an enormous amount of tracking data is generated daily in the domain of sports analytics. Vision-based player tracking systems have been deployed in professional basketball, soccer, baseball, tennis and cricket ~\cite{STATS, Prozone}. In multi-agent sports there exists several prediction opportunities, ranging from prediction of the occurrence of a single event, to prediction of a match outcome, or even the final score. However, instead of utilising raw trajectories directly for prediction tasks, current approaches ``hand-craft'' features and learn semantics at an abstract-level from these features ~\cite{Miller14, Yue14}. 

\par It is our observation that these approaches posses several inherent disadvantages compared to learning directly from raw trajectories: i) the generated feature representation is an overrepresentation which is inefficient in terms of memory\footnote{If a database consists of 10GB of raw-trajectories, but a feature representation is 10x or 100x this, then we will need to store 100GB or 1TB in memory which can be prohibitive.}, ii) the learnt feature representations are not interpretable and are non-informative for human operators interacting with the data, which limits potential fine-grain retrieval/knowledge-discovery applications \cite{fernando2015discovering}, and iii) the hand-crafted features may capture only abstract level semantics, which might lead to the loss of important semantics for the situation \cite{fernando2017soft+,fernando2017tree}. In literature such as \cite{Danyluk, Lee12, Schmidhuber,our_wacv,our_wacv2,fernando2017going,fernando2017Learning} the authors elaborate on the key advantages of moving away from hand-crafted solutions in favour of deep-learning approaches that learn features directly from the data.
\par In this paper, we propose a dictionary learning algorithm for adversarial multi-agent trajectories which  are encountered in competitive team sports. We are interested in predicting the occurrence of a goal from a shot in soccer. To the best of our knowledge this is the first work that addresses the above stated problem. The novelty of the proposed method is a deep decision tree based approach that automatically learns relevant features from multi-agent trajectories. Furthermore the learnt features are both interpretable for domain experts as well as informative for many short term prediction tasks. \par
In the proposed method, we first align the trajectories to a template which is learnt directly from the data, then we build our tree structure iteratively in a layer wise manner, alternating between  agglomerative clustering on feature weights and updating the weights using back propagation of classification error. The result is a ``deep decision" tree which ensures interpretability and the integrity of the feature representation learnt. The experimental results demonstrate the feasibility and applicability of the proposed method. 


\section{Related Work}

\subsection{Learning from handcrafted features}

Even in domains where large-amounts of data do exist such as sport, the first step has been to hand-craft features and then learn high-level features on top of these~\cite{Miller14, Yue14}. For example in ~\cite{Yue14}, the authors have learnt separate models for predicting the probability of passes, shots and ball possession by the current ball handler in basketball, and then combined these individual models to generate the final event prediction. 

 The main drawback of such handcrafted feature approaches is that complex dynamics cannot be modelled via a handful of features. Such features may capture only abstract concepts, and the learnt feature representation is non-interpretable; so it cannot be used for data-mining or knowledge discovery problems. 

\subsection{Learning from raw trajectories}
Recently, Lucey et al.~\cite{Lucey:2013} and Bialkowski et al.~\cite{Bialkowski14} used the raw trajectories to find soccer formations from player tracking data for visualisation and data clean-up, but this work did not consider prediction. Similarly, utilising video data from soccer games, the authors in ~\cite{Li:2009} have shown that group activity patterns can be derived from interactions among agents in the temporal domain. Cervone et al. ~\cite{Cervone2014} used basketball tracking data to predict player behaviour during a play. They used an expected possession value model that assumes that the decision of the ball-handler depends only on the current spatial configuration of the team in possession. Carr et al. ~\cite{Carr2013} used realtime player tracking data to predict the future location of play and point a robotic camera in that location for automatic sports broadcasting purposes. Yet none of these approaches have considered the problem of predicting future events. 

\subsection{Dictionary learning for trajectory data}
In the general application of dictionary learning to classification tasks, most previous approaches treat dictionary learning and classifier training as two separate processes, e.g. \cite{Boureau, Huang, Mairal, Grosse, Zhang}. In such approaches, a dictionary is learnt first and the subsequent dictionary representation is used to train a classifier such as an SVM. \par
In recent years supervised dictionary learning techniques have attracted much attention among the vision community. An iterative approach is proposed in \cite{Duc-Son}, which alternates between dictionary construction and classifier design. In \cite{conf/nips/MairalBPSZ08}, the authors have shown that this design may suffer from local minima. In \cite{YangYH10} the authors simultaneously learn an over-complete dictionary and multiple classification models for each class. However, the general assumption in the above mentioned dictionary learning is that each example is a linear combination of learnt dictionary elements. In the case of image patches this assumption may be valid, but in the case of dense non-liner trajectories this assumption is highly unlikely to hold. \par

Our dictionary learning approach is inspired by works from prediction works outside the multi-agent and trajectory literature. In particularly, we were motivated by the work of Hinton et al.~\cite{Hinton:2006} and Bengio et. al~\cite{Bengio07greedylayer-wise}. Hinton et al.~\cite{Hinton:2006} proposed a ``greedy learning algorithm for transforming representations'' to learn the weights of the deep belief network that classified the MNIST dataset. The weight vectors in each of the hidden layers were initialised greedily layer wise iteratively, and are updated using back propagation. In Bengio et. al~\cite{Bengio07greedylayer-wise}, the authors have shown that this achieves higher precision than random initialization of the weights. However a substantial amount of effort is required to extend the greedy layer wise learning concept to the multi-agent trajectory domain. 
Recently, Kontschieder et al. ~\cite{Kontschieder(2015)} proposed a Deep Neural Decision Forests structure which combines classification trees with unsupervised feature learning in convolutional neural networks. As described earlier, directly unsupervised feature learning techniques such as convolutional networks and deep belief nets ~\cite{Hinton:2006} will cause the loss of interpretability and integrity of the group representation of the multi agent trajectories. Therefore a considerable research gap exists when applying unsupervised feature learning and  deep learning techniques to model multi agent behaviour. 
\subsection{Match prediction in Soccer}
\par Considering the related work done in the area of soccer match score prediction,  Baio et al.~\cite{Baio:6677270} proposed a bayesian hierarchical model which incorporates home and away context into the prediction model. This was followed by the works by Owramipur et al. in ~\cite{rohtua:2013},  by Baker et al. in ~\cite{Baker2015} and Constantinou et al in ~\cite{Constantinou2012}. Yet these approaches \cite{rohtua:2013, Baker2015, Constantinou2012} have considered the problem of predicting the final results (i.e whether a win, loose or a draw) rather than predicting the final score. The latter is considered to be a more challenging problem as numerous factors have different degrees of influences when deciding the final score. 

\section{Methodology}

\subsection {Problem Definition}
We assume that each frame in our dataset is first preprocessed and we have obtained the spatial coordinates of each player at every time frame. Our observations of adversarial multi-agent behaviours come in the form of fine-grain spatial locations ($x,y$) of each agent sampled at a uniform rate (i.e., 10fps). The trajectory of the $i^{th}$ agent across a period of $\tau$ frames can then be defined by the vector, 
 \begin{equation}
\mathbf{x_{i}}= [x_{1},y_{1},\ldots,x_{\tau},y_{\tau}].
 \end{equation}

Given that we have a fixed number of agents, $m$, we can represent the behaviour of a group/team as a concatenation of each agent's trajectory,
  \begin{equation}
 \mathbf{X}_{A}=[\mathbf{x}^{T}_{1},\ldots,\mathbf{x}^{T}_{m}],
  \end{equation}
  where $A$ is the identity of the group. To incorporate an adversary, we concatenate one group with the second to form the ``play~representation'',
   \begin{equation}
  \mathbf{X}_{\mathrm{play}}=[\mathbf{X}_{A};\mathbf{X_{B}}].
   \end{equation}
\par Our goal is to learn a model which maps $\mathbf{X}_{\mathrm{play}}$ to the probability of a goal occurring from that particular shot, $P=[0, 1]$. To simplify the problem, all events are of fixed-length.  In the experiments $\tau$ is set to 100; and we select the 10 seconds (i.e. 100 entries at 10 fps) directly prior to the occurrence of the event of interest as our window for feature extraction. For simplicity we are considering only the player trajectory, however the ball trajectory could also be incorporated into the same framework. 

\par We have the additional constraint of dealing with a representation of the raw trajectories, so we have to ensure the integrity of the group representation (i.e., don't split up individual trajectories) as well as retaining interpretability. 
In this paper, our core contribution is the development of a framework to achieve this via a deep decision tree. Furthermore we demonstrate  this framework on applications of the learnt group attributes that are reliant on the preservation of interpretability. Due to the volume of data available, we focus on soccer tracking data, although our approach can be used in any adversarial multi-agent domain. 

\subsection{Data set}
We utilise player tracking data from 3 Premier League tournaments and 3 Champions League tournaments, obtained from Prozone and Amisco tracking systems, resulting in more than 2000 matches. The dataset provides player positions for every $1/10$ th of a second. Such fine grain precision is not available in any publicly available dataset.

\subsection{Automatic feature learning from trajectory data}
Discovering the playbook, or a set of plays which are representative of a team's behaviour is an unsupervised learning or clustering task. The simplest approach would be to only use the ball trajectory, and cluster plays accordingly. However, this is sub-optimal as the players and event information are extremely important and capture   semantics more effectively. Conversely, not all the players are involved in the play, making the dimensionality needlessly high.

\par Fig. 1 illustrates our approach. 
\begin{figure}[t]
\begin{center}
   \includegraphics[width=.99\linewidth]{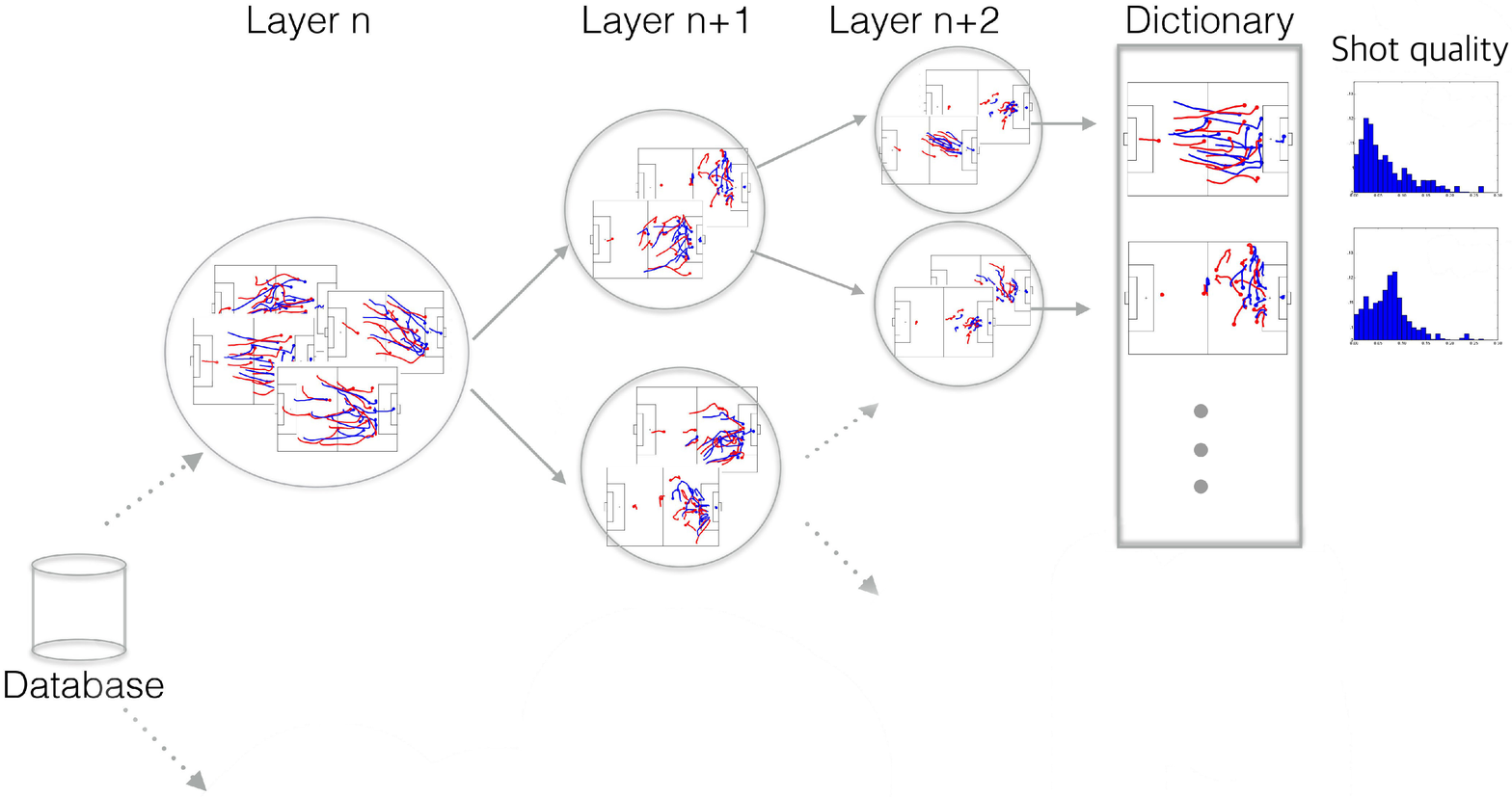}
\end{center}
\caption{Proposed Deep Decision Tree Architecture: Given a large database of tracking data, we first align the data to a formation template, and segment plays into coarse  groups. Then the decision tree with n+2 layers is formed. We use  back  propagation  to learn the tree which is semantically interpretable as well as informative to the prediction task at hand. Due to the interpretable and informative nature of this learnt dictionary, it can be used for many knowledge discovery tasks}
\label{fig:fig1}
\end{figure} 
Our proposed approach contains three main concepts: feature weights, decision nodes, and prediction nodes, defined as follows:

\begin{enumerate}
\item \textit{Feature weights} quantify the significance of each player trajectory in a given play.
\item \textit{Decision nodes} are internal nodes and are responsible for routing the samples through the tree structure using a specific decision function. The decision function used in the proposed work is a weighted clustering algorithm that utilises feature weights. 
\item \textit{Prediction nodes} are the leaf nodes in the tree. They generate the relevant classification using the assigned plays. 
\end{enumerate}

Given a player tracking  database, we first align the data to a formation template using the method proposed in ~\cite{Bialkowski14}. Aligning is necessary as the players tend to switch their positions throughout the game. Then these aligned plays are passed through a series of decision nodes which routes the plays based on the specified decision function which utilises feature weights for the routing process. When the samples reach the leaf nodes we generate relevant classifications using the prediction nodes. In the next step we back propagate this classification error and update the feature weights. This allows us to focus more on significant trajectories for the prediction task when routing the examples. This approach essentially is a decision-tree which is ``deep" in nature.

\par We compare our proposed approach with a Deep Convolution Neural Network (DCNN) approach, which is a bottom up approach. In a general DCNN application we start with raw pixels and we build features layer by layer, gathering more information. However this approach cannot be directly applied to multi agent trajectories due to the nature of multi agent behaviour. Any slight change in trajectory will cause another dictionary element to occur. Hence, every codebook element will have only 1 or 2 plays within each bucket; and the model will overfit.  DCNNs have found success with image and video data due to the enormous dimensionality; where variable imaging factors such as illumination and scale can render a top down approach infeasible. 
\par However, when considering the scenario of soccer, the problem is more constrained. The variation at the highest layer is much less than for an image problem. For instance, the number of players is almost always fixed, and the dimensionality is much smaller as we have only the player trajectories. Therefore we can reach the same solution in a much faster manner.  The proposed approach is outlined in the following subsections: Sections \ref{sec:dictionary} and \ref{sec:pred} explain the decision nodes (DN) and prediction nodes (PN) that make up the network. Section \ref{sec:backpropergation} explains how we use back propagation to learn feature weights within the network. The supplementary material presents the full algorithm for our proposed approach.

\subsubsection{Decision Nodes}
\label{sec:dictionary}

Decision nodes, $ b \ (b \in B $),  are internal nodes of the tree while prediction nodes, $ l \ (l \in L $), are the terminal nodes.  Each prediction node, $l \ $, holds a probability distribution  over $P$ (Figure 1 (a) far right). 
Each decision node $b$ is assigned a decision function which is responsible for routing samples along the tree.  The  decision  function is a feature weight based  clustering approach which we utilise to split the plays passed from the layer above. 
The reason for using agglomerative clustering based on classification instead of a general clustering approach is illustrated below. 

\par General clustering approaches aim to assign each example in the training set to a cluster which minimises the reconstruction error with respect to the number of clusters.
Let $\mathcal{X}=[\mathbf{X}^{\mathrm{play}}_1,\ldots,\mathbf{X}^{\mathrm{play}}_N]$ be the set of plays with offensive and defensive player 
trajectories. Learning a reconstructive dictionary with \( \vartheta \) items can be accomplished by,

 \begin{equation}
	 \sum\limits_{i=1}^\vartheta\sum\limits_{j}||\mathbf{X}^{\mathrm{play}}_j-\Upsilon_i||_2^2 \quad, 
 \end{equation}

where \(\Upsilon_i\) is the geometric centroid of the data points for cluster \(i\) and \(\vartheta\) is the number of clusters. The term \(|| \mathbf{X}^{\mathrm{play}}_j-\Upsilon_i||_2^2\) denotes the distortion. It should be noted that all the feature elements are weighted equally when evaluating the distortion measure. But for mutil-agent trajectories this is not optimal. Some trajectories are more significant than others for the prediction task that we are interested in. Therefore, we utilise a weighted distortion measure. 

Let $\mathbf{X}^{\mathrm{play}}_i=(\mathbf{x}_{\mathrm{i,1}},\ldots,\mathbf{x}_{\mathrm{i,m}}) \ $and $\mathbf{X}^{\mathrm{play}}_j=(\mathbf{x}_{\mathrm{j,1}},\ldots,\mathbf{x}_{\mathrm{j,m}})$ be two plays where $\mathrm{m}$ is the number of features (i.e $\mathrm{m}=22$ in the case of soccer as there are 22 player trajectories). $ \mathrm{\alpha}=(\mathrm{\alpha}_\mathrm{1}, \ldots, \mathrm{\alpha}_\mathrm{m})  $  
is the set of feature weights (we outline how these feature weights are learnt in Section  \ref{sec:backpropergation}). The weighted distortion between $\mathbf{X}^{\mathrm{play}}_i$ and $\mathbf{X}^{\mathrm{play}}_j$ is,
 \begin{equation}
\mathrm{D}^{\alpha}(\mathbf{X}^{\mathrm{play}}_i, \mathbf{X}^{\mathrm{play}}_j) =\sum\limits_{l=1}^\mathrm{m}\mathrm{\alpha}_{\mathrm{l}}
\mathrm{D}_{\mathrm{l}}(\mathbf{x}_{\mathrm{i,l}},\mathbf{x}_{\mathrm{j,l}}) ,
\label{eq:decsion}
\end{equation}
where $\mathrm{D}_{\mathrm{l}}(\mathbf{x}_{\mathrm{i,l}},\mathbf{x}_{\mathrm{j,l}})$, in the case of squared Euclidean distance is given by,
\begin{equation}
\mathrm{D}_{\mathrm{l}}(\mathrm{x}_{\mathrm{i,l}},\mathrm{x}_{\mathrm{j,l}})=(\mathrm{x}_{\mathrm{i,l}}-\mathrm{x}_{\mathrm{j,l}})^T(\mathrm{x}_{\mathrm{i,l}} - \mathrm{x}_{\mathrm{j,l}}).
\end{equation}

Therefore the objective of a decision node is to minimise Eq. \ref{eq:decsion}. By utilising a clustering objective as a decision function we are able to retain the semantic correspondence and the interpretability of the data provided. 


%

\subsubsection{Prediction Nodes (PNs)}
\label{sec:pred}
Once a sample ends in a leaf node, the related tree prediction is given by the classifier associated with  that particular leaf node. The passed player trajectories are the features for the classifier in the PNs. In addition to learning a reconstructive dictionary which learns different play representations, we are also interested in learning a dictionary that best discriminates the positive and negative examples for the prediction task that we are interested in. We incorporate this objective in prediction nodes as follows, 
\begin{equation}
||p_i-f(\mathbf{X}^{\mathrm{play}}_i,\mathbf{\pi}_k)||_2^2 \quad ,
\label{eq:dis_cluster}
\end{equation}
where $\mathbf{\pi}_k$ is the weight vector parameterising the classifier $f$ associated with the $k^{th}$ prediction node, and $\mathbf{p}=[p_1,\ldots, p_N]$ are the class labels associated with each play. The learnt probability distribution for each classifier can be utilised for many prediction tasks. A set of potential applications are demonstrated in Section \ref{sec:applications}.

\subsection{Back Propagation in the Decision Tree}
\label{sec:backpropergation}
In order to optimise the combined objective we can utilise the two step optimisation strategy given in \cite{Kontschieder(2015)} where the algorithm alternates the updates of $\alpha$ with updates of $\pi$. We start with segmenting plays based on semantics into free-kicks, open plays, corners and counter-attacks. Then we create the desired tree structure and randomly initialise the decision node parameters based on the procedure given in \cite{Kontschieder(2015)}. Parameters  $\alpha$ and  $\pi$ are learnt using stochastic gradient decent. 

%

\section{Evaluation}
In this section we evaluate the proposed prediction model and in Section \ref{sec:applications}, we demonstrate it's applications.

\subsection{Evaluation protocol}\label{eval1}
We randomly divided the dataset into training and testing sets where the training set contains 70\% of the matches (i.e 37,800 shots) and the testing set contains the remaining 30\% (i.e 16,200 shots). 
 As our baseline model we utilised the hand-crafted features proposed in ~\cite{Lucey:2013}.  i) the shot location, ii) positions of the closest 4 defenders, iii) positions of closest 4 attacking players, and iv) position of the goalkeeper. Using the above stated features we compared the accuracy of predicting the likelihood of occurrence of a goal from a shot using different classifiers: Random Decision Forests ~\cite{RandomForestHO}, Logistic Regression ~\cite{StatisticalModels:David2009} and SVMs ~\cite{Kecman2001}. 

\subsection{Evaluation results}
\begin{table}
\begin{center}
\begin{tabular}{|c | c | c|}
\hline
\textbf{No} & \textbf{Experiment} & \textbf{Mean Log Loss} \\
\hline\hline
\multirow{3}{*}{1} & Random Forest with 150 trees & 0.4913 \\
    & Logistic Regression & 0.9482 \\
    & SVM Regression & 0.9109 \\
\hline
2 & Decision tree with 2 layers & 0.3576  \\
\hline
3 & Decision tree with 4 layers & 0.0891\\
\hline
4 & Decision tree with 5 layers & 0.0889\\
\hline
5 & Decision tree with 6 layers & 0.0890\\
\hline \end{tabular}
\end{center}
\vspace{-5mm}
\caption{Evaluation results for short term prediction. As the baseline model we handcrafted features proposed in ~\cite{Lucey:2013} and trained different classifiers (row 1). Then we evaluated performance by altering the number of layers of the decision tree (rows 2 to 5).}\label{tab:tab0}
\end{table}
\vspace{-10mm}

\begin{figure}[h]
   \includegraphics[width=.8\linewidth]{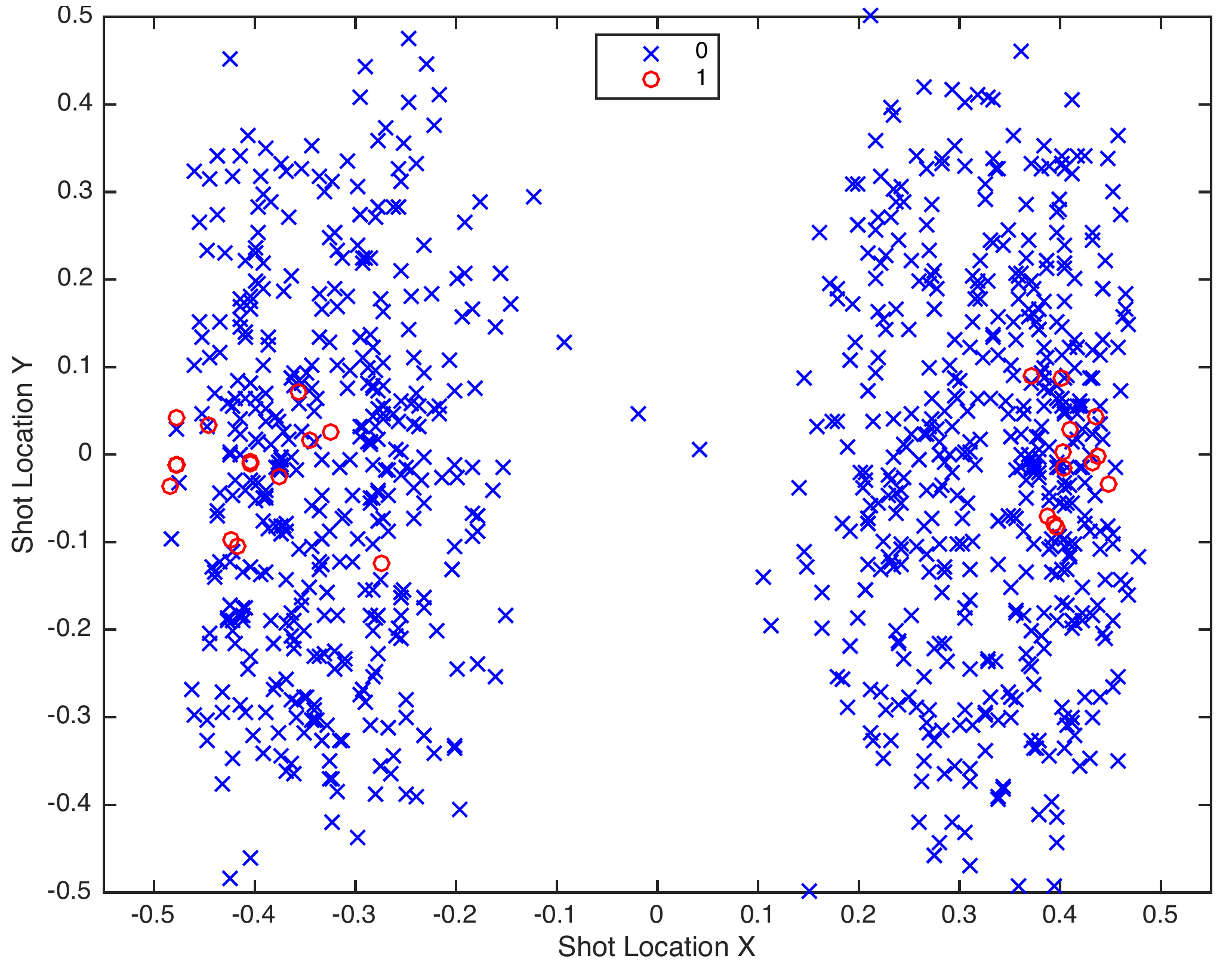}
   \caption{Distribution of handcrafted shot location feature in the first 1000 examples. It can be seen that is difficult to extract any high level understanding, or infer tactical information from the hand-crafted feature.}
   \label{fig:visualisation_of_hand_crafted_features}
\end{figure}

\begin{figure*}[!htb]
\begin{center}
   \includegraphics[width=1\textwidth]{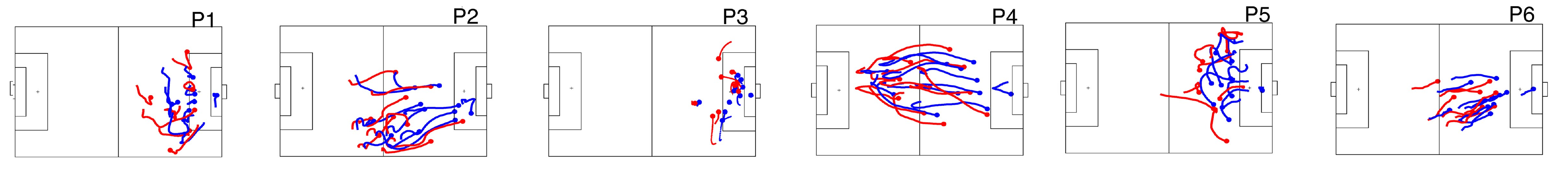}
\end{center}
\vspace{-10mm}
   \caption{Play book of scoring methods ( Red is attacking team running left-to-right. Blue is defensive team defending running right-to-left ). This figure shows only the first 6 of code-book elements. Please refer to the supplementary material for the complete codebook.}
\label{fig:all_clusters}
\end{figure*}

\begin{figure*}[!htb]
\begin{center}
   \includegraphics[width=1\textwidth]{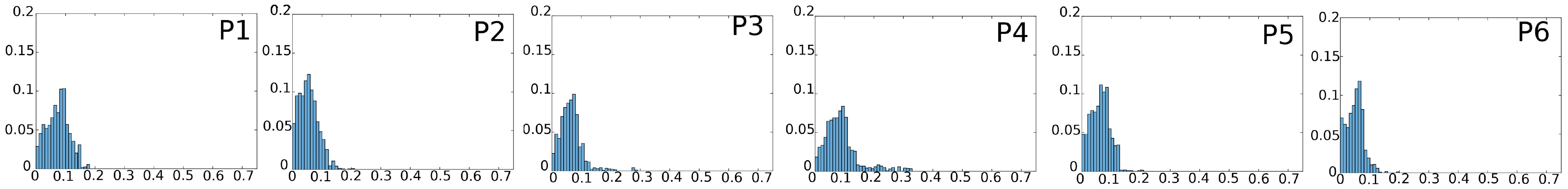}
\end{center}
\vspace{-10mm}
   \caption{Histogram of the expected goal values for each scoring method shown in Figure \ref{fig:all_clusters}. In all plots the x-axis show the expected goal value and the y-axis shows the frequency. Please refer to the supplementary material for the complete set of distributions.}
\label{fig:all_EGVs}
\end{figure*}

The prediction performance in terms of log-loss are shown in the second row of Table \ref{tab:tab0}.
Not surprisingly, the Random Decision Forest performed the best due to
its capacity to model complex behaviour. The optimal number of trees was chosen as 150 as this provided the minimum cumulative log-loss. Similarly, the maximum depth of each tree is set to 3 after evaluating trees of depth 2, 3, 5 and 10. 
Then utilising the proposed algorithm and extending the idea to  4 layers achieves a much better performance with a log-loss down to 0.0891. With the addition of a layer the log-loss is slightly reduced denoting that the model has converged with 4 layers (see rows 2-4). \par

An important feature of the proposed approach is that we maintain interpretability of the data. In Fig. \ref{fig:visualisation_of_hand_crafted_features} we have shown the distribution of the  shot location feature for the first 1000 examples in our database. We compare it with the learnt features of the proposed model (shown in Figure \ref{fig:all_clusters}). The proposed framework generates a codebook or a playbook of scoring methods that teams use for scoring goals, and an accompanying set of distributions quantifying the likelihood of a goal being scored from that particular codebook element (given in Figure \ref{fig:all_EGVs}).  The process of generating such distributions can be summarised as follows. \par
Given a set of bins $B_k= (l_k, \ldots, l_{k+1})$ with fixed bin width $h= l_{k+1}-l_k $, a set of plays $\mathcal{X}=[\mathbf{X}^{\mathrm{play}}_1,\ldots,\mathbf{X}^{\mathrm{play}}_N]$ and a set of codebook element $c_j= [c_1 \ldots c_{36}]$ , the decision tree assigns the given play $\mathbf{X}^{\mathrm{play}}_i$ to a respective codebook element $c_i$ and generates a probability $p_i$ of that play being a goal, 

\begin{equation}
(c_i, p_i)=f_{DecTree}(\mathbf{X}_{\mathrm{play}i}).
\end{equation} 

The bin counts for codebook type $c_{j}$ are given by,
\begin{equation}
v_{k,c_{j}}= \sum\limits^{N}_{i=1}I(c_{i} \in c_{j} \cap p_{i} \in B_k ),
\end{equation} 
where  $\sum\limits_{k}\sum\limits_{j}v_{k,c_{j}}=N$; and $I$ is an indicator function. 
Then the distribution for codebook type $c_{j}$ is given by,
\begin{equation}
f_{c_{j}}= {v_{k,c_{j}}\over{n_{c_{j}}h}} ,
\end{equation}

where 
\begin{equation}
n_{c_{j}}=\sum\limits^{N}_{i=1}I(c_{i} \in c_{j}) . 
\end{equation}
\par
Each of the codebook element captures the trajectories of all the players for a 10 second time window that leads to a shot on goal. It is worth noting that all shot types except penalties were included in this analysis. It can be seen from Figure \ref{fig:all_clusters}\footnote{Note that Figures 4 and 5 only show the first 6 of 36 plays in our playbook. Please see the supplementary material for for the complete codebook.} that our playbook captures several different plays such as counter-attacks (P4, down the centre, and P2, P6 down the right), and free-kicks (P3, P6).

Contrasting Figure \ref{fig:visualisation_of_hand_crafted_features} to Figure \ref{fig:all_clusters}, it is evident that handcrafted features don't generate a meaningful representation. Furthermore as they capture only abstract level semantics of the scene, for a single time instance they are not a rich information source for prediction. In contrast, our proposed unsupervised architecture captures the information for a 10 second time period leading to a shot on goal and outperforms the handcrafted approaches as shown in Table  \ref{tab:tab0}.

\begin{figure}[htb]
\begin{center}
\includegraphics[width=0.8\linewidth]{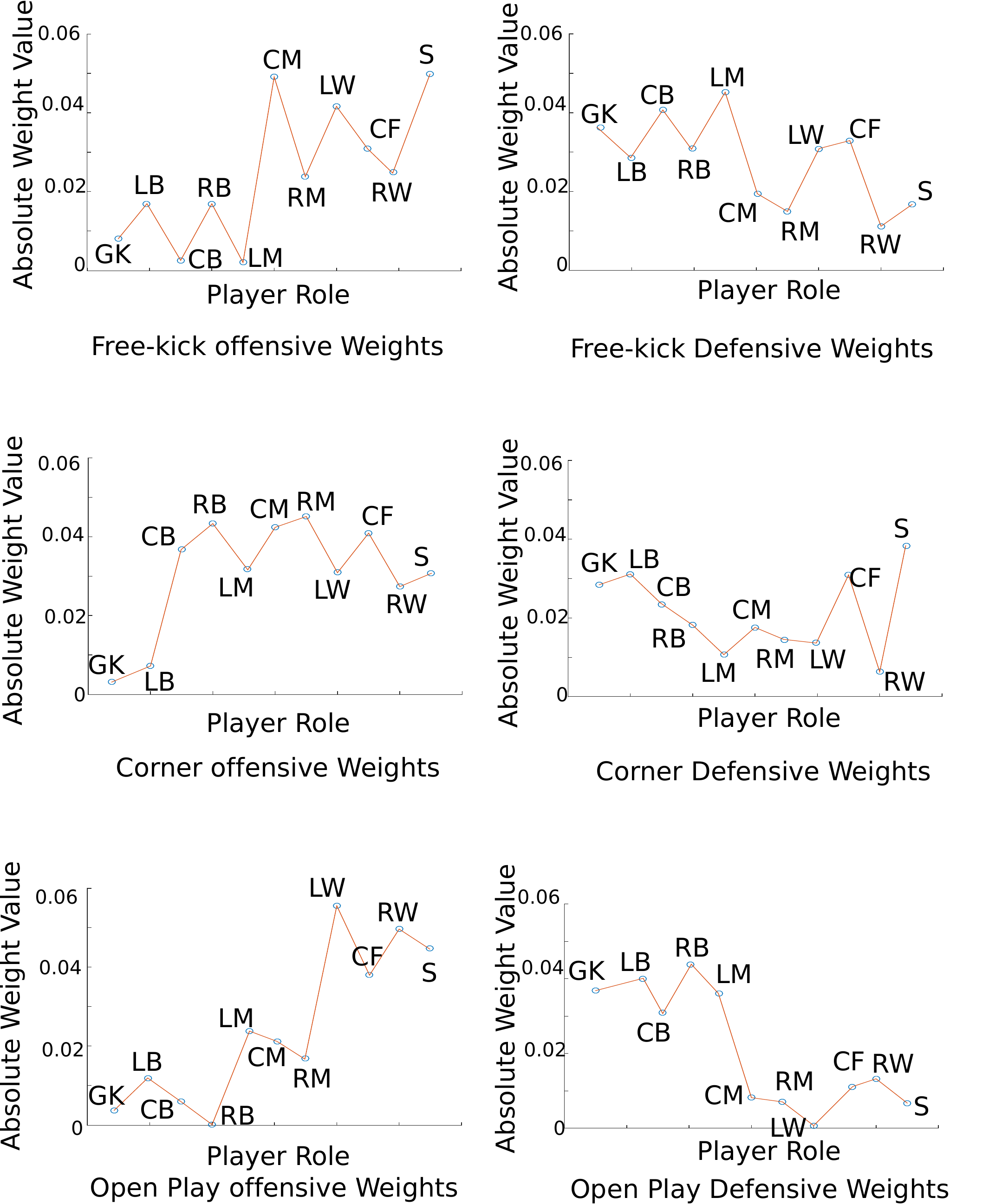}
\end{center}
\vspace{-5mm}
\caption{Absolute Weight Values $\alpha$, base on Role. It can be clearly seen that the importance of each role varies according to the type of play.}
\label{fig:fig3}
\end{figure}

Additionally with the proposed framework we are able to learn the weight of each role for offensive and defensive trajectories, under different contexts (Figure \ref{fig:fig3}); allowing us to quantify the importance of individual positions for different types of shots. This idea can be extended to each codebook element, where one could analyse what player roles influence the occurrence of a goal.

\section{Applications}
\label{sec:applications}

\subsection{Team strategy analysis}
As illustrated earlier, our proposed approach provides us with a codebook of scoring methods for soccer and likelihood distributions specifying the quality of that shot type (i.e likelihood of a goal being scored from that shot type). If we analyse this at a team level, and assign the shots played by a particular team to the closest codebook element that we have and then quantify it's quality (i.e. was a goal scored), we can represent the strategy of a particular team.\par
\begin{figure}[htb]
\begin{center}
   \includegraphics[width=1.0\linewidth]{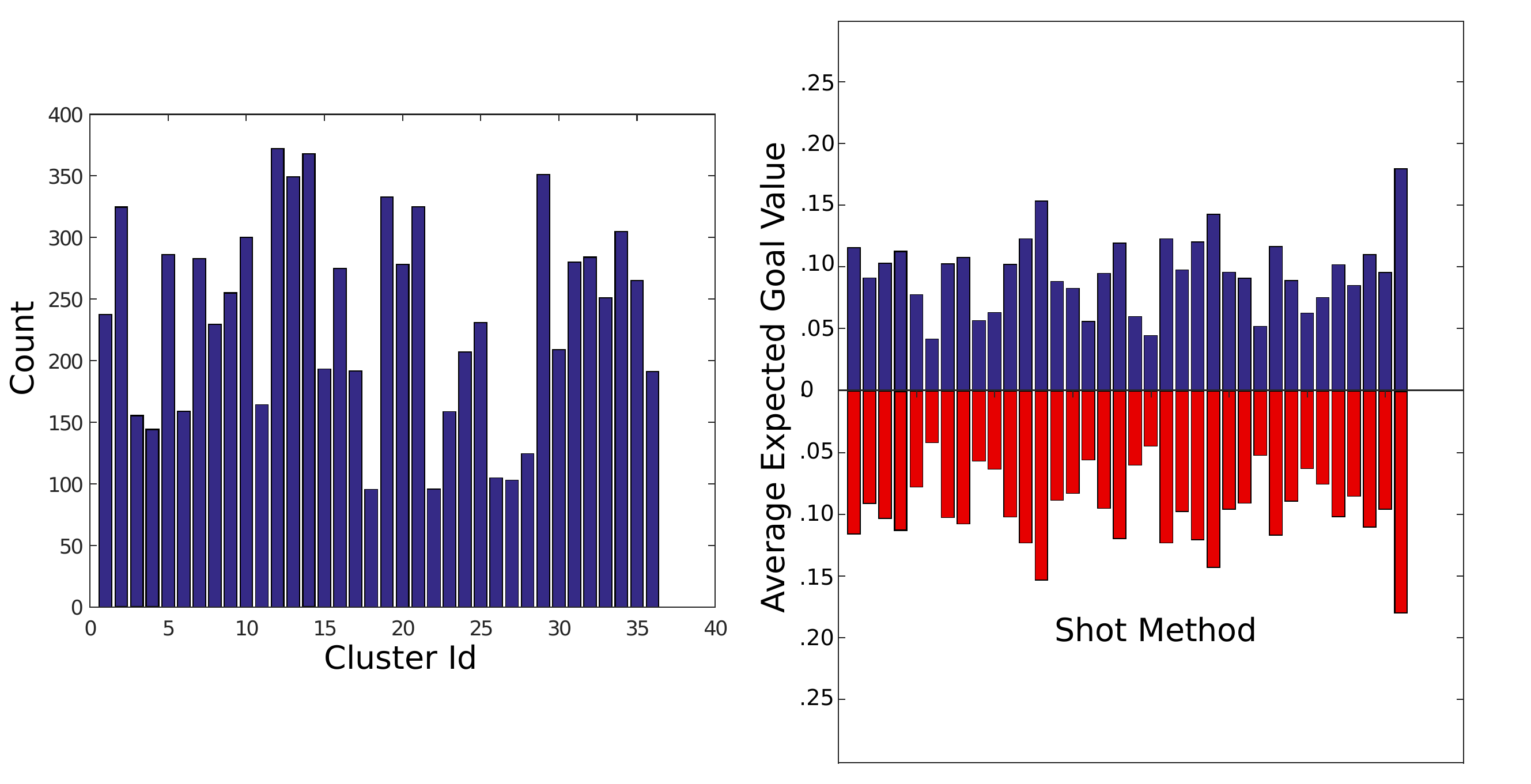}
\end{center}
\vspace{-8mm}
   \caption{The left plot shows the frequency of each shot type, P12 and P14 are the most common. The right plot shows the mean "Strategy Plot", which shows the expected goals per method offensively (blue) and defensively (red). We can use this to normalise an individual teams distribution to understand how they compare to the league average. }
\label{fig:all_cluster_counts}
\end{figure}

With that intent, from a recent season of a top-tier European league (380 games, 38 per team, 19 home and 19 away matches), we analysed at a team level which methods teams use to create chances offensively, as well as concede defensively. \footnote{Due to privacy constraints, we have  anonymised the league and the team identities. As such we identify the teams as letters ranging from A-T.} 
Figure \ref{fig:all_cluster_counts} shows league wide tendencies for scoring and conceding. From Figure \ref{fig:all_cluster_counts} (a), we can see that P12 and P14 are the most common methods of creating a shot. But in terms of effectiveness, we can see from Figure \ref{fig:all_cluster_counts} (b) that there exists vast variability in the effectiveness of shooting methods. This strategy plot is calculated by obtaining the mean expected goal value per shot method. This gives us a goal value per shooting method and visualise of whether one team converts or concedes a chance in this manner. \par
Given the set of plays $\mathbf{X}^{\mathrm{play}}_i=[\mathbf{X}^{\mathrm{play}}_1,\ldots,\mathbf{X}^{\mathrm{play}}_N]$ in the season, the mean offensive strategy distribution can be generated by, 
\begin{equation}
f^{MSO}=\sum\limits_{j}v_{c_{j}}^{*} ,
\end{equation}
where
\begin{equation}
v_{c_{j}}^{*}={\sum\limits_{i=1}^{N}I(c_{i} \in c_{j})q_{i}\over{n_{c_{j}}}} .  
\end{equation}

\par
\begin{figure}[!htb]
\begin{center}
   \includegraphics[width=1\linewidth]{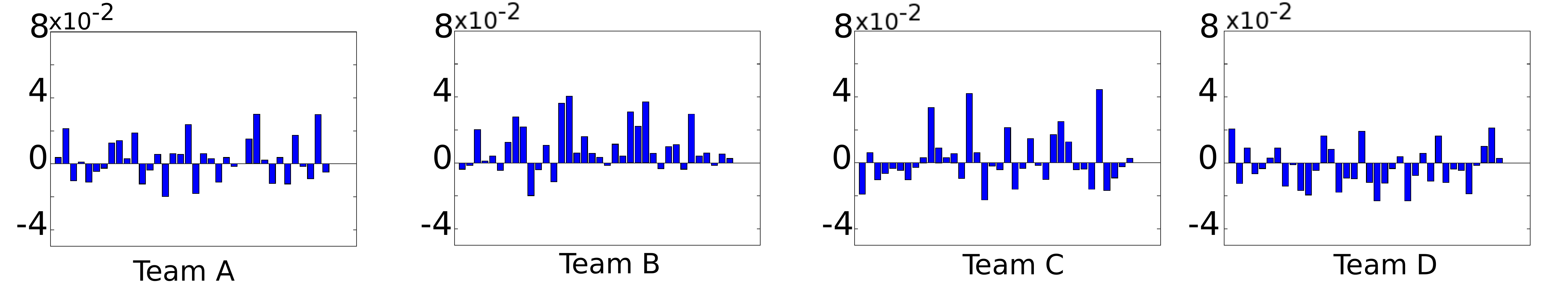}
\end{center}
\vspace{-8mm}
   \caption{Relative Offensive Strategy Plot. League wide offensive strategy is subtracted at a team level. In all plots the x-axis shows the shot type and y-axis shows the $f^{RSO}$.}
\label{fig:relative_offense}
\end{figure}

\begin{figure}[!htb]
\begin{center}
   \includegraphics[width=1\linewidth]{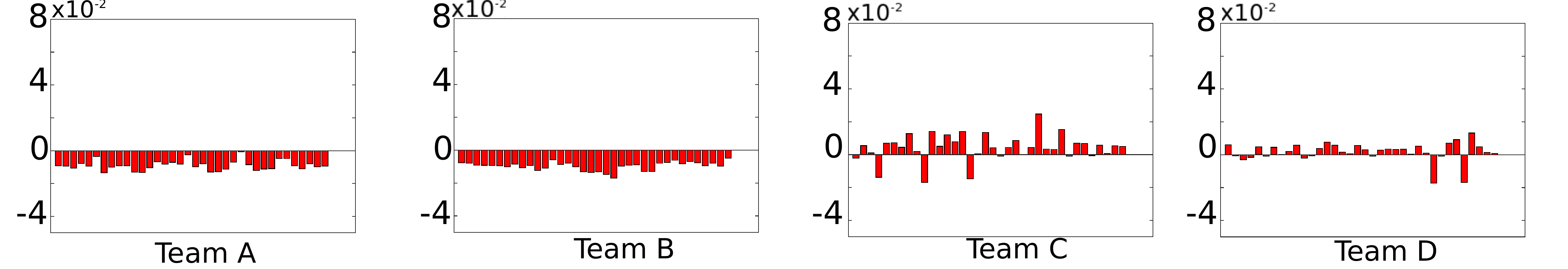}
\end{center}
\vspace{-8mm}
   \caption{Relative Defensive Strategy Plot. League wide average defensive strategy is subtracted at a team level. In all plots the x-axis shows the shot type and y-axis shows the $f^{RSD}$.}
\label{fig:relative_defense}
\end{figure}

 Even though the average strategy plot is quite uniform, we can see in Figures \ref{fig:relative_offense} and \ref{fig:relative_defense}, at a per team level  that they are highly variable. We generate the relative strategy plots (Figures \ref{fig:relative_offense} and \ref{fig:relative_defense}) by subtracting the league average from the team's strategy plot.
The offensive strategy of team $t$ where $T_k=[\mathrm{Team_1}, \ldots, \mathrm{Team_{22}}]$ is given by,
\begin{equation}
f^{TSO}_{t}=\sum\limits_{j}v_{(c_{j}, t)}^{*} ,
\end{equation}
where
\begin{equation}
v_{(c_{j},t)}^{*}={\sum\limits_{i=1}^{N}I(c_{i} \in c_{j} \cap t \in T_k)q_{i}\over{n_{(c_{j},t)}}} . 
\end{equation}
In the above equation $n_{(c_{j},t)}$ is given by
\begin{equation}
n_{(c_{j},t)}=\sum\limits^{N}_{i=1}I(c_{i} \in c_{j} \cap t \in T_k) . 
\end{equation}
Then we can generate the relative strategy offence of team $t$ by,
\begin{equation}
f^{RSO}_{t}=f^{TSO}_{t}-f^{MSO} .
\end{equation}

In Fig. \ref{fig:relative_offense} we have the relative offensive capability, compared to Fig. \ref{fig:relative_defense} where we show the defensive capability per shooting method. Note that in Fig. \ref{fig:relative_offense} and Fig. \ref{fig:relative_defense} we have only the first 4 teams (A-D)\footnote{Please refer to the supplementary material for complete set of plots}.
For the offensive plots, having more positive bins suggests that teams are offensively stronger compared to the league average, where  specific peaks or dips highlight a particular strength or deficiency respectively for that particular scoring method. 

\par For the defensive plots, the opposite applies; having more positive peeks suggests that the team is susceptible to that scoring method, while more bars below suggests that that team is strong in not allowing that particular scoring method to lead to a goal compared to the league-average. From these plots we can gain insights into a teams strengths and weaknesses. For instance, we can see that while Team A is inconsistent in attack, they are strong in defence; while team C is very effective for a number of attacking plays, they are also more vulnerable than average in defence.

\subsection{Predicting Long Term Behaviour: Match Simulation and Betting}
Long-term events can be viewed as an aggregation of short-term events. By utilising the features that improved the short term prediction one can also improve long-term prediction.
The idea behind soccer match simulation is to predict when and how (i.e which shot type) teams are going to shoot, and how many shots will result in goals. As we have distributions quantifying the probability of each team scoring a goal from each shot type (shown in Fig. \ref{fig:relative_offense}), and the defensive strength of each team against different shot types (shown in Fig. \ref{fig:relative_defense}), we can aggregate the final score.\par
 A problem with this naive approach is that the probability of an event is dependent on the context, such as the score, home team, and time remaining.
 As such, the prediction of the long-term outcome needs to incorporate this context. This approach is in contrast to current long term event modelling approaches where the sub-event likelihoods are fixed for the entire event.  In this section, we show that by incorporating the context and learning a model for this situation we can improve long-term prediction.

An approach for soccer match simulation is proposed by Baio et al.~\cite{Baio:6677270}, which incorporates home and away context.
The number of goals scored by the home and the away team in the
$\mathbf{g^{th}}$ game of the season is given by $\mathbf{y_{g1}}$
and $\mathbf{y_{g2}}$ respectively. The vector of observed counts,
$\mathbf{y = (y_{g1}, y_{g2})}$ is modelled as an independent Poisson:
\begin{center}
\(y_{gj} |\theta_{gj} \sim Poisson(\theta_{gj} ) \)
\end{center}
where the parameters,
\begin{center}
\(\theta= (\theta_{g1}, \theta_{g2}),\)
\end{center}
represent the shooting intensity in the g-th game for the team playing
at $\mathbf{home (j = 1)}$ and $\mathbf{away (j = 2)}$.

The random variable $\mathbf{\theta}$ can be modelled using a
log-linear random effect model,
\begin{center}
\(log\theta_{g1} = home + att_{h(g)} + def_{a(g)}\)
\(log\theta_{g2} = att_a(g) + def_{h(g)}.\)
\end{center}

The parameter $\mathbf{home}$ represents the advantage for the team
hosting the game and we assume that this effect is constant for all
teams throughout the season.

However, as described previously these descriptions are rather coarse and do not give specifics about a team's ability in different contexts. For example, a team maybe more or less susceptible to ``corner kicks'' or ``counter attacks'' depending on their style of play. Therefore when playing against such a team an opponent may vary their style. Furthermore a team may decide to attack or defend depending on the current score and the time remaining. 
%
\subsubsection{Simulating Different Contexts}

The decision tree indicates the likelihood of an event
occurring under different contexts. As such, simulation
allows us to estimate what the cumulative effects will be on long-term
prediction as we sequentially go through events (e.g. what happens if
a team scoring early vs a team scores late).

Our baseline model (\textbf{M1}) is to fix the probabilities of an event
occurring. At the start of the simulation the clock is set to 00:00
and the occurrence of the next shot by the two teams is predicted via linear regression. The current time is selected and the probability
of the event (i.e., goal being scored) is quantised to either 0 or 1
by randomly sampling the shot distribution of each team. The current
score and time is updated accordingly. The process is repeated until
the current time exceeds the stoppage time. We ran every  simulation
1000 times and the average score is recorded. We then used our method (\textbf{Our}), in which the linear regression model predicting the time of the occurrence of the next shot takes context features into account. The selected context features are: a binary feature indicating whether its home/away; the current score and the remaining time.

\subsubsection{Evaluation protocol}
Similar to Sec. \ref{eval1}, we divide the dataset into training and testing sets, where the training set contains 70\% of matches in the original dataset and the testing set contains the rest. We evaluated the average Mean Squared Error (MSE) for 3 prediction models: the Bayesian hierarchical model \textbf{(BHM)} of \cite{Baio:6677270}; our baseline model (\textbf{M1}); and our proposed method (\textbf{Our}). The results are presented in Table \ref{tab:table_0}.
\subsubsection{Match simulation results}
The results indicate that our approach outperforms the BHM and M1 due to the fact that it allows the model to capture the contextual information. 
 When comparing the MSE generated from individual models, M1 has the highest MSE as it doesn't include any contextual information. The prediction process goes blindly with a fixed probability distribution for each team. When moving to the BHM model the MSE decreases as it takes home/ away context into account. With our model we show that one can further improve the prediction process through an advanced contextual model which incorporates a rich set of features. 
\begin{table}
\begin{center}
 \begin{tabular}{|c|c|c|}
 \hline
\textbf{Exp. No}&\textbf{Method}&\textbf{MSE}\\
\hline
1&BHM: Bayesian hierarchical model \cite{Baio:6677270}&1.92\\
\hline
2&M1: Fixed likelihood model &2.15\\
\hline
3&Ours: Varying likelihood model & \textbf{1.42}\\
\hline
\end{tabular}
\end{center}
\vspace{-5mm}
\caption{Evaluation results for match simulation: Mean Square Error (MSE) is shown for the BHM \cite{Baio:6677270}, M1 and our proposed model.}\label{tab:table_0}
\end{table}

\section{Conclusion}
In this paper, we have proposed a novel dictionary learning algorithm for adversarial multi agent sports tracking data. The proposed algorithm performs unsupervised learning of group attributes directly from tracking data. The discriminative nature of the algorithm ensures that learnt codebook elements are ideal for short term prediction tasks. Furthermore the proposed deep decision tree architecture preserves the interpretability of the learnt codebook, enabling domain experts to perform high level data mining and knowledge discovery activities on the learnt dictionary.  The experimental results indicate a vast range of applications for these codebook plays.  We demonstrated the applicability of this technique  not only for short term predictions such as predicting the occurrence of an event, but also for long term predictions such as simulating games between two opponents. The informative nature of the codebook also allows for strategic analysis  permitting us to quantifying and compare different strategies among teams.

{\small
\bibliographystyle{ieee}
\bibliography{my_references}

\begin{thebibliography}{10}\itemsep=-1pt

\bibitem{Baio:6677270}
G.~Baio and M.~A. Blangiardo.
\newblock {Bayesian hierarchical model for the prediction of football results}.
\newblock pages 1--13, 2010.

\bibitem{Baker2015}
R.~D. Baker and I.~G. McHale.
\newblock Time varying ratings in association football: the all-time greatest
  team is..
\newblock {\em Journal of the Royal Statistical Society: Series A (Statistics
  in Society)}, 178(2):481--492, 2015.

\bibitem{Bengio07greedylayer-wise}
Y.~Bengio, P.~Lamblin, D.~Popovici, H.~Larochelle, U.~D. MontrŽal, and
  M.~QuŽbec.
\newblock Greedy layer-wise training of deep networks.
\newblock In {\em In NIPS}, pages 153--160. MIT Press, 2007.

\bibitem{Bialkowski14}
A.~Bialkowski, P.~Lucey, P.~Carr, Y.~Yue, and I.~Matthews.
\newblock Large-scale analysis of soccer matches using spatiotemporal tracking
  data.
\newblock In {\em ICDM}, pages 725 -- 730, 2014.

\bibitem{Boureau}
Y.-L. Boureau, F.~R. Bach, Y.~LeCun, and J.~Ponce.
\newblock Learning mid-level features for recognition.
\newblock In {\em CVPR}, pages 2559--2566. IEEE Computer Society, 2010.

\bibitem{Constantinou2012}
A.~C. Constantinou, N.~E. Fenton, and M.~Neil.
\newblock pi-football: A bayesian network model for forecasting association
  football match outcomes.
\newblock {\em Knowledge-Based Systems}, 36:322 -- 339, 2012.

\bibitem{Cervone2014}
L.~B. D.~Cervone, A.~DÕAmour and K.~Goldsberry.
\newblock Predicting points and valuing decisions in real time with nba optical
  tracking data.
\newblock {\em MIT Sloan Sports Analytics Conference,}, 2014.

\bibitem{fernando2017tree}
T.~Fernando, S.~Denman, A.~McFadyen, S.~Sridharan, and C.~Fookes.
\newblock Tree memory networks for modelling long-term temporal dependencies.
\newblock {\em arXiv preprint arXiv:1703.04706}, 2017.

\bibitem{fernando2017going}
T.~Fernando, S.~Denman, S.~Sridharan, and C.~Fookes.
\newblock Going deeper: Autonomous steering with neural memory networks.
\newblock In {\em International Conference on Computer Vision Workshops
  (ICCVW)}, pages 214--221, 2017.

\bibitem{fernando2017soft+}
T.~Fernando, S.~Denman, S.~Sridharan, and C.~Fookes.
\newblock Soft+ hardwired attention: An lstm framework for human trajectory
  prediction and abnormal event detection.
\newblock {\em arXiv preprint arXiv:1702.05552}, 2017.

\bibitem{fernando2017Learning}
T.~Fernando, S.~Denman, S.~Sridharan, and C.~Fookes.
\newblock Learning temporal strategic relationships using generative
  adversarial imitation learning.
\newblock {\em International Foundation for Autonomous Agents and Multiagent
  Systems}, 2018.

\bibitem{our_wacv}
T.~Fernando, S.~Denman, S.~Sridharan, and C.~Fookes.
\newblock Task specific visual saliency prediction with memory augmented
  conditional generative adversarial networks.
\newblock {\em Winter Conference on Applications of Computer Vision (WACV)},
  2018.

\bibitem{our_wacv2}
T.~Fernando, S.~Denman, S.~Sridharan, and C.~Fookes.
\newblock Tracking by prediction: A deep generative model for multi-person
  localisation and tracking.
\newblock {\em Winter Conference on Applications of Computer Vision (WACV)},
  2018.

\bibitem{fernando2015discovering}
T.~Fernando, X.~Wei, C.~Fookes, S.~Sridharan, and P.~Lucey.
\newblock Discovering methods of scoring in soccer using tracking data.
\newblock {\em Large-Scale Sports Analytics, Sidney}, 2015.

\bibitem{StatisticalModels:David2009}
D.~A. Freedman.
\newblock Statistical models: Theory and practice.
\newblock {\em Cambridge University Press}, pages 128--130, 2009.

\bibitem{Grosse}
R.~Grosse, R.~Raina, H.~Kwong, and A.~Y. Ng.
\newblock Shift-invariant sparse coding for audio classification.
\newblock In {\em Proceedings of the Twenty-third Conference on Uncertainty in
  Artificial Intelligence}, 2007.

\bibitem{Hinton:2006}
G.~E. Hinton, S.~Osindero, and Y.-W. Teh.
\newblock A fast learning algorithm for deep belief nets.
\newblock {\em Neural Comput.}, 18(7):1527--1554, July 2006.

\bibitem{RandomForestHO}
T.~K. Ho.
\newblock Random decision forests.
\newblock In {\em Proceedings of the Third International Conference on Document
  Analysis and Recognition (Volume 1) - Volume 1}, ICDAR '95, pages 278--,
  Washington, DC, USA, 1995. IEEE Computer Society.

\bibitem{Huang}
K.~Huang and S.~Aviyente.
\newblock Sparse representation for signal classification.
\newblock In B.~Schšlkopf, J.~C. Platt, and T.~Hoffman, editors, {\em NIPS},
  pages 609--616. MIT Press, 2006.

\bibitem{Kecman2001}
V.~Kecman.
\newblock Learning and soft computing Ñ support vector machines, neural
  networks, fuzzy logic systems.
\newblock {\em The MIT Press, Cambridge, MA,}, 2001.

\bibitem{Kontschieder(2015)}
P.~Kontschieder, M.~Fiterau, A.~Criminisi, and S.~R. Bulo'.
\newblock Deep neural decision forests. [winner of the david marr prize 2015].
\newblock In {\em Intl. Conf. on Computer Vision (ICCV), Santiago, Chile},
  December 2015.

\bibitem{Danyluk}
H.~Lee, R.~B. Grosse, R.~Ranganath, and A.~Y. Ng.
\newblock Convolutional deep belief networks for scalable unsupervised learning
  of hierarchical representations.
\newblock In A.~P. Danyluk, L.~Bottou, and M.~L. Littman, editors, {\em ICML},
  volume 382 of {\em ACM International Conference Proceeding Series}, page~77.
  ACM, 2009.

\bibitem{Lee12}
H.~Lee, P.~T. Pham, Y.~Largman, and A.~Y. Ng.
\newblock Unsupervised feature learning for audio classification using
  convolutional deep belief networks.
\newblock In Y.~Bengio, D.~Schuurmans, J.~D. Lafferty, C.~K.~I. Williams, and
  A.~Culotta, editors, {\em NIPS}, pages 1096--1104. Curran Associates, Inc.,
  2009.

\bibitem{Li:2009}
R.~Li, R.~Chellappa, and S.~K. Zhou.
\newblock Learning multi-modal densities on discriminative temporal interaction
  manifold for group activity recognition.
\newblock In {\em CVPR}, pages 2450--2457. IEEE Computer Society, 2009.

\bibitem{Lucey:2013}
P.~Lucey, D.~Oliver, P.~Carr, J.~Roth, and I.~Matthews.
\newblock Assessing team strategy using spatiotemporal data.
\newblock In {\em Proceedings of the 19th ACM SIGKDD International Conference
  on Knowledge Discovery and Data Mining}, KDD '13, pages 1366--1374, New York,
  NY, USA, 2013. ACM.

\bibitem{Mairal}
J.~Mairal, F.~Bach, J.~Ponce, G.~Sapiro, and A.~Zisserman.
\newblock Discriminative learned dictionaries for local image analysis.
\newblock In {\em CVPR}, pages 1 -- 8. IEEE Computer Society, 2008.

\bibitem{conf/nips/MairalBPSZ08}
J.~Mairal, F.~R. Bach, J.~Ponce, G.~Sapiro, and A.~Zisserman.
\newblock Supervised dictionary learning.
\newblock In D.~Koller, D.~Schuurmans, Y.~Bengio, and L.~Bottou, editors, {\em
  NIPS}, pages 1033--1040. Curran Associates, Inc., 2008.

\bibitem{Miller14}
A.~Miller, L.~Bornn, R.~Adams, and K.~Goldsberry.
\newblock {Factorized Point Process Intensities: A Spatial Analysis of
  Professional Basketball}.
\newblock In {\em ICML}, pages 235Ð--243, 2014.

\bibitem{rohtua:2013}
F.~Owramipur, P.~Eskandarian, and F.~S. Mozneb.
\newblock Football result prediction with bayesian network in spanish
  league-barcelona team.
\newblock {\em International Journal of Computer Theory and Engineering},
  5(5):812, 10 2013.

\bibitem{Carr2013}
M.~M. P.~Carr and I.~Matthews.
\newblock Hybrid robotic/virtual pan-tilt-zoom cameras for autonomous event
  recording.
\newblock {\em ACM Multimedia,}, 2013.

\bibitem{Duc-Son}
D.-S. Pham and S.~Venkatesh.
\newblock Joint learning and dictionary construction for pattern recognition.
\newblock {\em CVPR}, pages pp. 1 -- 8, 2008.

\bibitem{Prozone}
{Prozone}.
\newblock \url{http://www.prozonesports.com/index.html}.

\bibitem{Schmidhuber}
J.~Schmidhuber.
\newblock Deep learning in neural networks: An overview.
\newblock {\em Neural Networks}, 61:85--117, 2015.
\newblock Published online 2014; based on TR arXiv:1404.7828 [cs.NE].

\bibitem{STATS}
{Stats}.
\newblock \url{http://www.stats.com/}.

\bibitem{YangYH10}
J.~Yang, K.~Yu, and T.~S. Huang.
\newblock Supervised translation-invariant sparse coding.
\newblock In {\em CVPR}, pages 3517--3524. IEEE Computer Society, 2010.

\bibitem{Yue14}
Y.~Yue, P.~Lucey, P.~Carr, A.~Bialkowski, and I.~Matthews.
\newblock {Learning Fine-Grained Spatial Models for Dynamic Sports Play
  Prediction}.
\newblock In {\em ICDM}, pages 670 -- 679, 2014.

\bibitem{Zhang}
W.~Zhang, A.~Surve, X.~Fern, and T.~Dietterich.
\newblock {Learning non-redundant codebooks for classifying complex objects}.
\newblock In {\em ICML '09: Proceedings of the 26th Annual International
  Conference on Machine Learning}, pages 1241--1248, New York, NY, USA, 2009.
  ACM.

\end{thebibliography}
}

\end{document}